\documentclass[conference]{IEEEtran}
\IEEEoverridecommandlockouts

\usepackage{amsmath,amssymb,amsfonts}
\usepackage{algorithmic}
\usepackage{graphicx}
\usepackage{textcomp}
\usepackage{xcolor}
\usepackage{hyperref}
\usepackage{orcidlink}
\usepackage{siunitx}
\usepackage{booktabs} 
\usepackage{todonotes}
\usepackage{threeparttable}
\usepackage{float}
\usepackage{subfig}
\usepackage{multirow}
\usepackage[numbers]{natbib}
\begin{document}

\title{Image Compression with Bubble-Aware Frame Rate Adaptation for Energy-Efficient Video Capsule Endoscopy








\thanks{$^1$University of T\"ubingen, Faculty of Science, Department of Computer Science, Embedded Systems Group,
\tt\small {[first].[lastname]@uni-tuebingen.de}}
\thanks{This work has been partly funded by the German Federal Ministry of Research, Technology and Space (BMFTR) in the projects MEDGE (16ME0530) and Scale4Edge (16ME012).}}

\author{\IEEEauthorblockN{Oliver Bause$^1$}
\IEEEauthorblockA{
\orcidlink{0009-0003-5388-2959}}
\and
\IEEEauthorblockN{Jörg Gamerdinger$^1$}
\IEEEauthorblockA{
\orcidlink{0009-0006-8433-3292}}
\and
\IEEEauthorblockN{Julia Werner$^1$}
\IEEEauthorblockA{
\orcidlink{0009-0006-0279-1776}}
\and
\IEEEauthorblockN{Oliver Bringmann$^1$}
\IEEEauthorblockA{
\orcidlink{0000-0002-1615-507X}}
}

\maketitle

\begin{abstract}
Video Capsule Endoscopy (VCE) is a promising method for improving the medical examination of the small intestine in the gastrointestinal tract. 
A key challenge is their limited size, resulting in a short battery lifetime which conflicts with high energy consumption for image capturing and transmission to an on-body device. 
Thus, we propose an image compression pipeline that substantially reduces the transmitted data while preserving diagnostic image quality. 
Furthermore, we exploit characteristics of the compression process to identify frames with low diagnostic value mainly caused by bubbles, without requiring additional image analysis. 
For low-visibility frames, a dynamic bubble-aware frame rate adaptation strategy reduces image acquisition and transmission during these phases while preserving sensitivity to potential anomalies.
The proposed compression and frame rate adaptation are evaluated on a RISC-V platform using the Kvasir-Capsule and Galar datasets. 
The compression method achieves a compression ratio of $5.748$ $(82.6\%)$ at a peak signal-to-noise ratio of $\qty{40.3}{\dB}$, indicating negligible loss of visual quality.
The compression accomplished a mean energy reduction of the whole system by $20.58\%$.
Additionally, the proposed bubble-aware frame rate adaptation reduced the energy consumption by up to $40\%$. 
These results demonstrate the potential of our method to increase the applicability of VCE.
\end{abstract}


\section{Introduction}

Wireless Video Capsule Endoscopy (VCE) is a medical procedure that was first introduced in the early 2000s and involves a pill-sized capsule swallowed by a patient~\cite{iddan2000wireless,swain2001wireless}. 
The capsule is equipped with a miniature image sensor, LEDs for illumination, a transmitter unit, a battery pack, and a microcontroller and can be extended by additional sensors for increased diagnostic opportunities.
The device traverses the gastrointestinal (GI) tract, captures images, and sends them to an outside on-body receiver to be later evaluated by medical experts.
The procedure aims to detect pathologies with focus on the small intestine as this area is not accessible by the gastroscopy or colonoscopy~\cite{costamagna2002prospective}.
However, a major limitation of the capsule is the battery capacity and thus the runtime, as the available space within the device is severely restricted in order to keep it swallowable.
For example, for the PillCam™ SB3 from Medtronic an operation time of 8 to 12 h is reported with 2 to 6 frames per second (fps)~\cite{pillcam}.
The time required for the video capsule (VC) to traverse the GI tract, varies significantly among patients and may extend beyond the 12-hour time frame in certain cases.
As a result, screening of the colon or the small intestine will be incomplete, and potential pathologies may not be captured.

\textbf{Our Contribution:}
\begin{figure}
    \centering
    \includegraphics[width=\linewidth]{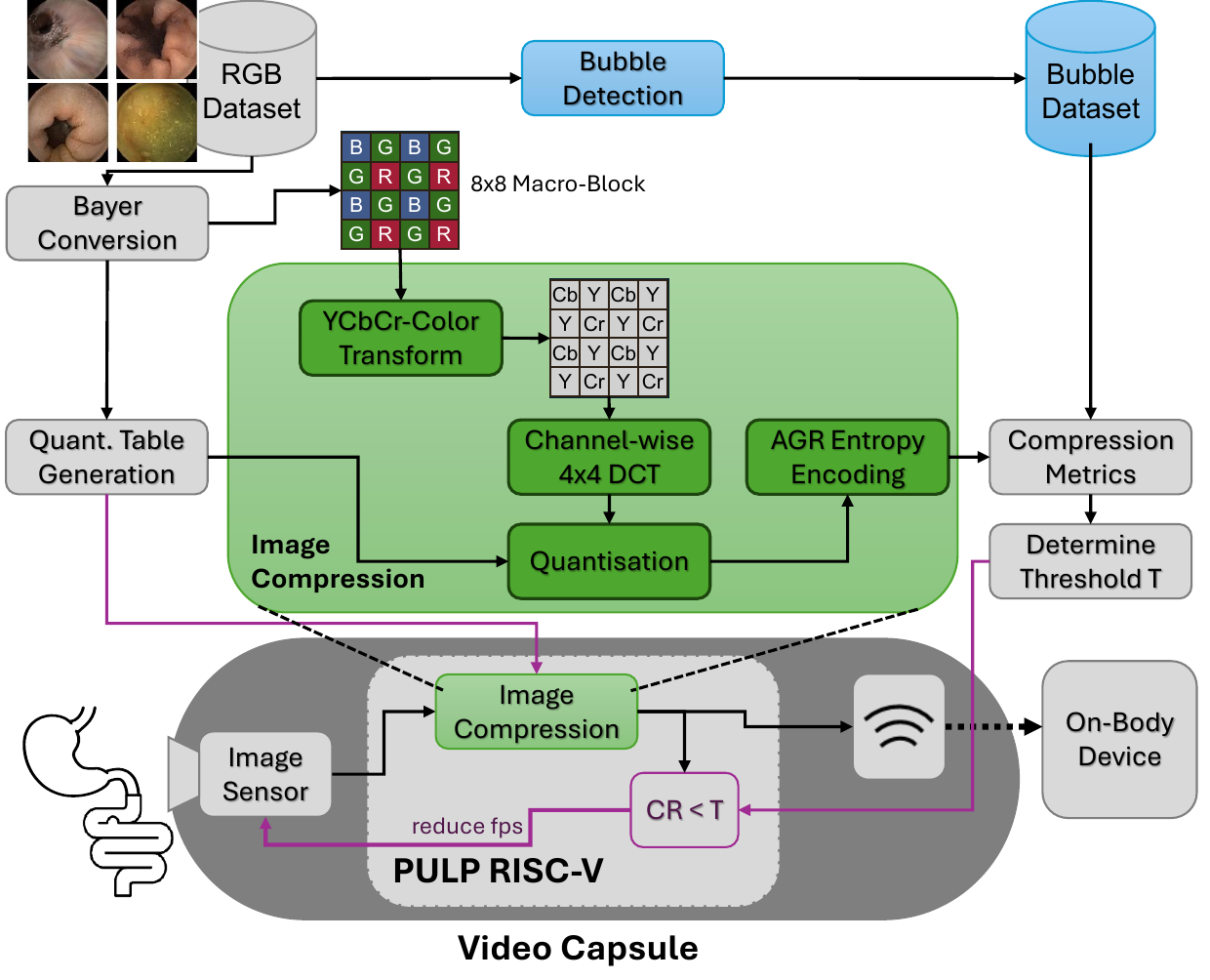}
    \caption{Proposed method of the RAW image compressing pipeline in combination with a newly generated bubble dataset to determine image visibility on-capsule.}
    \label{fig:pipeline}
\end{figure}
To prolong the operation time without increasing the battery capacity of VCs, this work aims to reduce the power consumed per image captured and transmitted as well as utilizing the available energy in a targeted manor.
The proposed pipeline is illustrated in Figure ~\ref{fig:pipeline}.
First, we introduce a hardware-suitable adaptive Golomb-Rice (AGR) coding compression technique~\cite{rice1971coding, malvar2006adaptive} that is compatible with the RAW Bayer output of the image sensor.
Since VCEs are equipped with a low resolution sensors, the compression should exhibit minimal loss in order to prevent large deterioration in quality.
This Rice coding reduces the amount of bits to be transmitted per frame which is the second largest energy consumer after the image capturing.

The prevailing generation of VCs does not analyze the frames on-device and merely transmits each captured image directly to the on-body receiver.
Nevertheless, certain images are not pertinent to the evaluation. 
This is due to the fact that they were either captured outside the area of interest (i.e., the small intestine) or are covered in small bubbles and dirt, thereby obscuring the intestinal wall.
The latter can also be addressed by our AGR coding as the compressed size negatively correlates with the area covered by bubbles.
Therefore, if the captured image can not be effectively compressed, the frame rate will be reduced until the visibility increases or rather compression ratio (CR) decreases again.
Thus, transmitting less irrelevant frames and withholding the energy for more relevant sections.
The process is validated with the datasets Kvasir-Capsule~\cite{smedsrud2021kvasir} and Galar~\cite{lefloch2025galar}, and demonstrated with a ultra-low power RISC-V System-on-Chip (SoC) demonstrator.
\section{Related Work}

\subsection{VCE Image Compression}
Previous work on VCE image compression has explored a range of techniques balancing CR, image quality (peak signal-to-noise ratio (PSNR)), power consumption, and inference time~\cite{sushma2022recent}.
~\cite{mostafa2014improved}'s improved YEF-DCT compressor uses an integer DCT with a custom color space to achieve an average compression rate of about $85 \%$ and a high PSNR of around $\qty{52}{\dB}$, focusing on low complexity suited for capsule integration.
However, their pipeline uses RGB images as input which needs demosaicking of the RAW sensor output and increases memory requirements threefold.
~\cite{lin2006ultra}'s and ~\cite{khan2011lossless}'s custom hardware image compression accelerator (ACC) target energy efficiency, reducing video size by $\geq75\%$ with a PSNR of $\qty{32.5}{\dB}$ and $72\%$ with a lossless compression, respectively.
~\cite{lin2006ultra}'s consumes roughly $\qty{14.9}{\milli\W}$ at 2 fps which is too much for a targeted battery life of more than $\qty{12}{\hour}$, while ~\cite{khan2011lossless}'s only requires $\qty{287}{\uW}$ at the same frame rate.
However, both are custom-made chips that are expensive to manufacture at small scale and need to be additionally integrated into the capsule besides the micro-controller that is handling the system.
~\cite{turcza2013hardware}’s hardware-efficient system integrates a DCT-based compressor with on-chip processing, achieving a PSNR metric of $\approx\qty{33}{\dB}$ with lower power consumption of $\qty{7}{\milli\W}$ in a $\qty{65}{\nano\metre}$ FPGA prototype with an inference time of $\qty{8.2}{\milli\s}$ per frame.

More recent learning-based compression by ~\cite{harshitha2025energy} demonstrates that a CNN-based feature learning algorithm can improve performance metrics marginally, reporting $\approx0.28\%$ higher compression ratio and $\approx0.15\%$ higher PSNR.
However the reported power consumption of $\qty{2.5}{\W}$ exceeds the VCE specifications by a factor of exponential magnitude with regard to achieving the requisite battery life.
Additionally, the memory demands of the utilized ResNet50 model (up to $\qty{100}{\mega\byte}$) are not suitable for ultra-low power devices which often have less than $\qty{2}{\mega\byte}$ of total memory capacity.

\subsection{Bubble Detection and Segmentation}
\label{subsec:bubble_detection}
Bubble detection and segmentation has been studied to mitigate visual artifacts that obscure diagnostically relevant content.
~\cite{bashar2010automatic} addressed this problem indirectly through automatic detection of informative frames, where frames heavily affected by bubbles, debris, or blur are identified as non-informative and filtered out using handcrafted features and classical classifiers. 
~\cite{segui2012categorization} further investigated intestinal content categorization and segmentation, proposing methods using Gabor filters and SURF detectors to distinguish bubbles from other content such as food residues and fluids, enabling more robust frame-level analysis.

More focused approaches explicitly target bubble structures. 
~\cite{mir2024identification} proposed a method for identifying circular patterns characteristic of bubble frames, leveraging geometric and shape-based features to detect bubbles effectively. 
~\cite{sadeghi2023segmentation} advanced this line of work by performing detailed segmentation and region quantification of bubbles in small bowel capsule endoscopy images using wavelet-based texture analysis, allowing precise localization and measurement of bubble regions. 
Both relied on Hough transforms~\cite{Hough1962method} to detect the circular bubbles on the Kvasir-Capsule dataset~\cite{smedsrud2021kvasir} which does not offer a ground truth labeling of bubble masks, which they generated separately.

\subsection{Real-Time Adjustment of VCE Sensor Configuration}
Dynamic system control has been proposed to improve battery efficiency in capsule endoscopy by adapting operational parameters to patient-specific conditions.
~\cite{sahafi2022edge} introduced an edge-AI–based capsule system that adjusts acquisition settings on-device, most notably dynamically changing the fps based on real-time image analysis to lower fps in less informative areas and increase it when relevant content is detected.
However, their early prototype lacks optimization, resulting in a reported battery life of just 1h.

~\cite{bause2025smartVCE} further advanced this idea by using lightweight on-capsule AI on RAW Bayer images for organ localization to explicitly control system behavior. 
Their method adapts the frame rate and activates image transmission only after detecting entry into the area of interest, the small intestine, significantly reducing energy usage before the region of interest is reached.
This patient-adaptive fps control accounts for large inter-patient variability in gastrointestinal transit times and enables substantial energy savings without compromising examination completeness.
Though, it requires a highly customized SoC with a dedicated ultra-low power hardware CNN-accelerator

From a clinical standpoint, ~\cite{yung2016capsule} highlighted that modern capsule colonoscopy systems increasingly rely on such adaptive strategies, including variable frame rates depending on capsule motion and anatomical location, to balance image quality, coverage, and battery lifetime. Collectively, these works demonstrate that dynamic adjustment of the capsule's sensor settings, with focus on frame rate, based on patient-specific and real-time information is a key mechanism for efficient and reliable capsule endoscopy.
\section{Methodology}
Based on this previous research, we aim to combine the on-capsule image compression with a visibility evaluation to leverage the power reduction twice.
First, we save energy by reducing the transmission size of each frame with a compression pipeline tailored to ultra-low power SoCs without the need for expensive customized hardware.
Second, a visibility evaluation is conducted by evaluating the compression rate of the latest frame, and, if the visibility is reduced by bubbles, the frame rate will be reduced until it increases again.
This is then demonstrated with a single-core RISC-V Pulpissimo SoC~\cite{bernardo2024scalable} without the utilization of the instruction set extension.

\subsection{Capsule Datasets}
\label{subsec:datasets}
The VCE datasets Kvasir-Capsule~\cite{smedsrud2021kvasir} and Galar~\cite{lefloch2025galar} were used to develop and validate the proposed framework.
These include relevant labeling regarding the visibility quality of frames.
Kvasir-Capsule is restricted to images of the small intestine and introduces the label \textit{reduced mucosal view} for images with content like stool or blood.
Galar is the currently largest available VCE dataset with 80 complete studies.
Moreover, it provides a technical label group for six studies that centers on image quality.
Firstly, frames of selected studies are annotated with one of the three labels \textit{good view}, \textit{reduced view}, or \textit{no view} which indicates a visibility reduction of less than $50\%$, between $50$ and $95\%$, and over $95\%$, respectively.
In addition, a distinction is made between whether the reduction is caused by bubbles or dirt.
However, both datasets do not contain any labeling of the number, positioning, nor dimensions of bubbles.

\subsection{Bayer Compression Pipeline}
The design and implementation of the lightweight image compression pipeline optimized for resource-constrained embedded platforms is shown in Figure ~\ref{fig:pipeline}.
The system processes RAW Bayer image data through a multi-stage pipeline consisting of color space transformation, frequency decomposition, quantization, and adaptive entropy coding.
The majority of image sensors output a frame with a $2\times2$ Bayer color filter array applied, where each pixel is assigned a dedicated color, with two green, one blue, and one red pixel per block.
To obtain the 3-channel RGB image, demosaicking is necessary.
However, this additional step would increase the latency and power consumption of the compression pipeline without providing any discernible benefit.
As illustrated in Figure ~\ref{fig:pipeline}, the RAW sensor output is utilized as input directly.

\subsubsection{Encoding}
The compression architecture is designed to minimize memory footprint and computational overhead while maintaining high reconstruction quality. 
The pipeline operates on $8 \times 8$ pixel macro-blocks, sub-divided into $4 \times 4$ blocks for frequency analysis.

\subsubsection{Reversible Color Transformation and Channel Separation}
To exploit inter-channel redundancy within the Bayer pattern, the RAW input is first transformed into a luminance-chrominance space. 
Unlike standard RGB-to-YUV transforms that require floating-point arithmetic, we implement a modified Reversible Color Transform (RCT) using integer shift-and-add operations.
The transform maps RAW Bayer samples (B, G1, G2, and R) into components that reduce spectral correlation. 
For a $2 \times 2$ Bayer cell, the transformation is defined to capture luminance ($Y$) and color differences ($C_b, C_r$):
\begin{itemize}
    \item \textbf{Luminance ($Y$):} Derived from the green channels to capture structural intensity.
    \item \textbf{Chrominance ($C_b, C_r$):} Derived from color differences, often centered at an offset (e.g., 128) to maintain unsigned 8-bit consistency.
\end{itemize} 
Following the transform, samples are reorganized into discrete $4 \times 4$ blocks, facilitating independent frequency processing for luminance and chrominance data.

\subsubsection{Fixed-Point $4\times4$ Discrete Cosine Transform}
To achieve energy compaction, we employ a $4\times4$ Discrete Cosine Transform (DCT). 
To meet the constraints of the embedded processor, the DCT is implemented using a high-speed, fixed-point integer approximation.
The row-column decomposition method is utilized to transform spatial residuals into frequency coefficients:
\begin{equation}
X_k=\sum_{n=0}^{N-1} x_n \cos{\left[\frac{\pi}{N} (n + \frac{1}{2})k\right]},
\label{Xk_eq}
\end{equation}
where $X_k$ is the DCT coefficient, $x_n$ is the input pixel, $N$ is the number of samples ($N=4$ for a $4\times4$ DCT), $n$ is the spatial index, and $k$ is the frequency index where a higher $k$ indicates higher spatial frequencies.
The 2D DCT can be decomposed into two 1D transforms. 
The implementation utilizes a hard-coded $4\times4$ kernel with scaled integer constants $S_{fixed}$, so the irrational constants are moved to a post-processing scaling stage to allow the core transform to use only computational simpler integer additions and shifts. 
The scaling matrix $S_{fixed}$ represents the normalization factors for the $4 \times 4$ basis functions:
\begin{itemize}
    \item Base Scale ($2^{12}$): $4096$
    \item Orthogonality Factor ($\sqrt{2} \cdot 2^{12}$): $\approx 5793$
    \item Combined Factor ($2 \cdot 2^{12}$): $8192$
\end{itemize}
This ensures that the $(i, j)$ coefficient is scaled by $k(i) \cdot k(j)$, where $k(0)=1$ and $k(k \ne 0)=\sqrt{2}$, effectively mapping the transform to a fixed-point domain without a Floating Point Unit (FPU).
Additionally, a Zero-AC fast path optimization is integrated.
If all AC coefficients in a block are zero, the encoder skips the frequency scan and entropy coding for those coefficients, significantly reducing cycles during processing of low-detail areas.
Finally, the DCT results are quantized by a channel-specific quantization table that was derived by the unquantized DCT statistics of the images of the Galar dataset.
The tables were limited to power-of-two values to only perform shift instead of expensive division operations.

\subsubsection{Zigzag Ordering and Entropy Coding}
The DCT coefficients are reordered using a zigzag scan to prioritize low-frequency components and maximize zero-value runs.
The final stage employs AGR coding~\cite{malvar2006adaptive}, an entropy coding scheme suited for embedded systems as it avoids complex probability tables.
Further optimizations include an encoder that maintains a running context for DC and AC values to adapt to local image statistics.
The optimal Rice parameter $k$ is calculated dynamically using the accumulated magnitude of previous symbols.
And finally, a \textit{FastBitWriter} manages the bit-level concatenation into 32-bit words, ensuring memory alignment and reducing expensive memory accesses.

\subsection{Bubble Dataset}
\begin{figure}[tb]
    \centering
    \includegraphics[width=\linewidth]{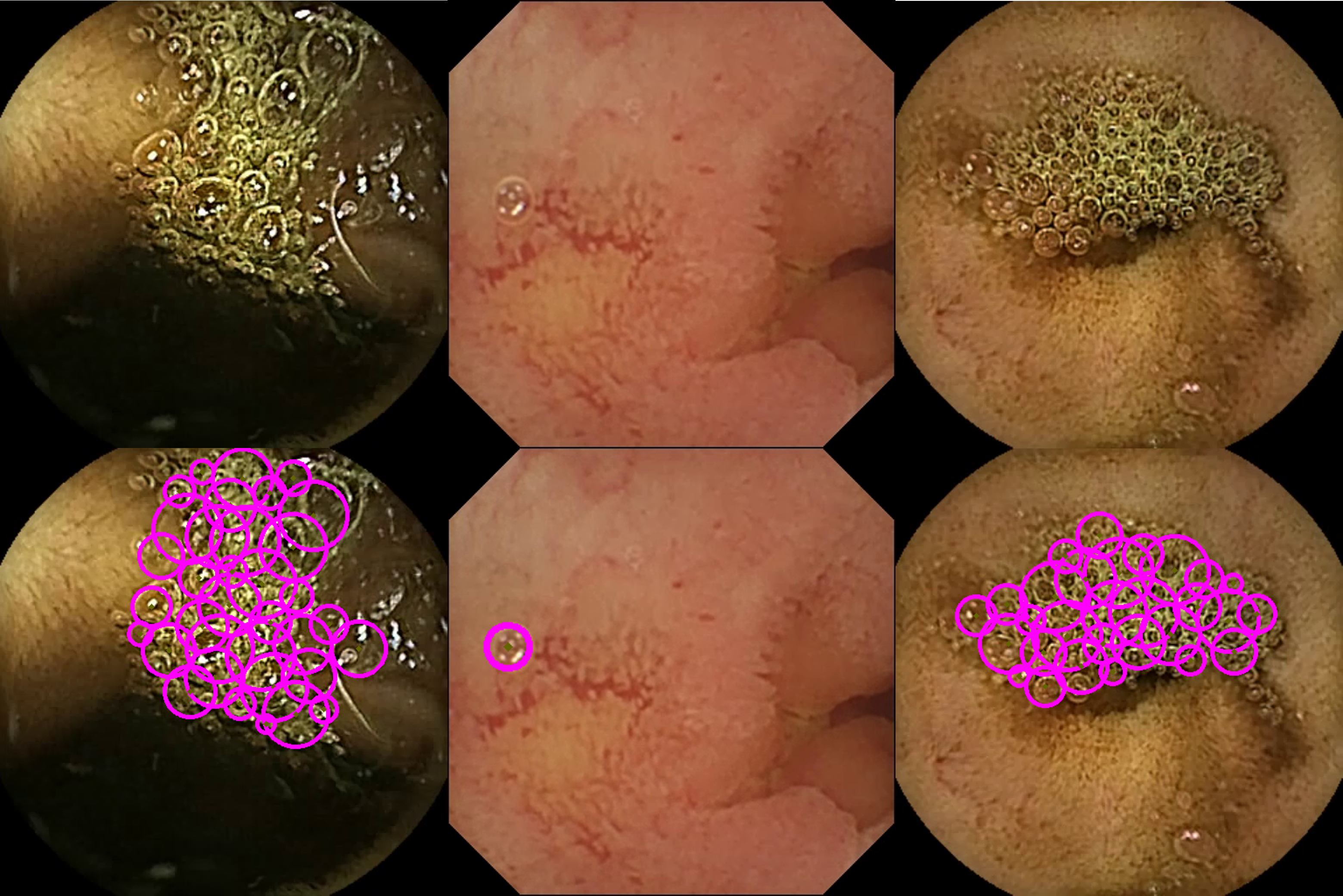}
    \caption{Example images from Kvasir-Caspule and Galar dataset (top) and the corresponding images with detected bubbles (bottom).}
    \label{fig:bubble-detection-example}
\end{figure}

As shown in Section~\ref{subsec:datasets} Kvasir-Capsule and Galar lack information about the number, positions and dimensions of bubbles within the dataset. Hence, we performed a post-processing on those datasets to improve their labeling using a Hough transformation which has been explored as well suited for bubble detection (see Section~\ref{subsec:bubble_detection}). Examples of the post-processing are shown in Figure~\ref{fig:bubble-detection-example}, an overview of the resulting datasets is shown in Table~\ref{tab:dataset_bubbles}.

The Hough transformation~\cite{Hough1962method} is a robust method for detecting lines or circles in images. It is based on gradient images, which are the binary result of edge detection. To this end, the image is first converted into an 8-bit grayscale image. Then, a median blur with a kernel size of  $5\times 5$ is applied, since the Hough transformation produces fewer false positives for smoothed images. As the appearance of circles varies slightly depending on the task, optimising the Hough transformation parameters is necessary to reliably detect bubbles. We derived improved parameters for both datasets using parameter space exploration. Since most of the bubbles are small, we limited the circle radius to a range of 3–30 px. The minimum distance between the centres of two bubbles is set to 10 px to reduce false behaviour in bubble clusters. In relation to the objective of this study, it is more important to determine the area covered by the bubbles than to identify each individual bubble within a cluster. The Canny threshold for edge detection is set to 100, representing the threshold at which a pixel value is classified as an edge pixel and incorporated into the binary edge mask. A lower threshold leads to multiple false positives, while a higher threshold limits correct edge classification since bubble edges appear smooth in images. The accumulator threshold, representing the required confidence during the detection stage to count a circle as a true positive, is 23. For the inverse ratio of the accumulator to the image, parameter space exploration led to a value of 0.9. This value is suitable, as lower values of about 1.0 achieve higher precision, and the increased memory consumption can be disregarded, since this is only used for offline labelling post-processing on a GPU server.

\begin{table}[]
\centering
\caption{Bubble Detection Results}
\begin{threeparttable}
\begin{tabular}{lcc}
\toprule
Dataset & Kvasir-Caspule~\cite{smedsrud2021kvasir} & Galar~\cite{lefloch2025galar} \\ \midrule
Frames & 47,238 & 1,375,918 \\
Frames with Bubbles & 18,810 (39.82\%) & 869,147 (63.17\%) \\
Bubbles Detected & 256,227 & 15,742,534 \\
Mean Coverage\tnote{*} {[}\%{]} & 13.14 & 14.86 \\
Median Coverage\tnote{*} {[}\%{]} & 5.04 & 7.21 \\
STD\tnote{*} {[}\%{]} & 0.183 & 0.185 \\ 
Mean Bubble Radius {[}px{]} & 22.38 & 24.15 \\
Median Bubble Radius {[}px{]} & 21 & 23 \\\bottomrule
\end{tabular}
\begin{tablenotes}
        \item[*] Only including images with at least one bubble detected.
    \end{tablenotes}
    \end{threeparttable}
    \label{tab:dataset_bubbles}
\end{table}

Table ~\ref{tab:dataset_bubbles} reports the resulting post-processed Kvasir-Capsule~\cite{smedsrud2021kvasir} and Galar~\cite{lefloch2025galar} datasets including the number of bubbles detected and their coverage.
The Galar dataset was constraint to the small intestine, as this region is the area of interest.
Galar includes whole studies, as opposed to the pre-selection of images of Kvasir-Capsule, which results in a significantly higher percentage of images with bubbles, reaching $63.17\%$.
However, a high degree of similarity exists between the distribution of the coverage area and bubble size in both datasets.
\section{Results}

\subsection{Metrics Definition}
Following metrics are typically used to evaluate the quality of compression algorithms.
\textbf{Compression Ratio (CR)} quantifies the effectiveness of a compression method by comparing the original data size to the compressed data size. 
It is defined as
\begin{equation}
    CR=\frac{S_{original}}{S_{compressed}},
\label{CR_eq}
\end{equation}
where $S_{original}$ is the size of the uncompressed data and $S_{compressed}$ is the size after compression. 
A larger CR indicates better compression efficiency.

\textbf{Peak Signal-to-Noise Ratio (PSNR)} measures the reconstruction quality of an image by comparing it to the original image and is derived from the Mean Squared Error (MSE):
\begin{equation}
    MSE = \frac{1}{MN}\sum_{i=1}^M\sum_{j=1}^N\left(I(i,j) - \hat{I}(i,j)\right)^2
\label{MSE_eq}
\end{equation}
\begin{equation}
    PSNR = 10\log_{10}\left(\frac{MAX^2_I}{MSE}\right),
\label{PSNR_eq}
\end{equation}
where $I$ and $\hat{I}$ are the original and reconstructed images, $M \times N$ is the image resolution, and $MAX_I$ is the maximum possible pixel value, e.g. 255 for 8-bit images. 
Higher PSNR indicates a less lossy compression, whereas a lossless compression would achieve a PSNR of $\infty$.

\textbf{Inference time (IT)} represents the computational time required to process a single input sample.
For real-time operation, the inference time must satisfy that the inference time is smaller than the maximum allowable time per input determined by the application’s real-time constraints.

\subsection{Compression on the Datasets}
\begin{figure}[t]
    \centering
    \subfloat[Kvasir-Capsule]{\includegraphics[width=.5\linewidth]{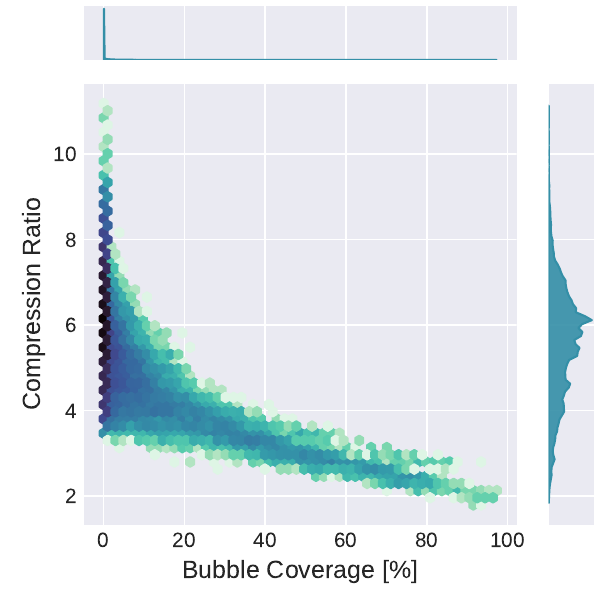}\label{fig:kvasir_comp_ratio}}
    \subfloat[Galar]{\includegraphics[width=.5\linewidth]{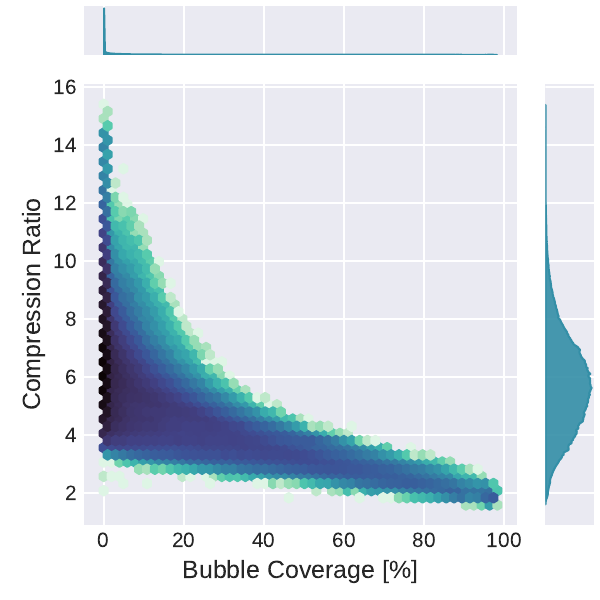}\label{fig:galar_comp_ratio}}
    \caption{Relation between the CR of the Bayer images and the bubble coverage with images of (a) Kvasir-Capsule and (b) Galar (small intestine only).}
    \label{fig:comp_ratio}
\end{figure}

All VCE datasets only contain the post-processed and demosaicked RGB images from the capsules.
These were subsampled by applying the RGGB color filter array to obtain a Bayer image as close to the sensor's original one as possible.
Then, both datasets were compressed by the proposed pipeline (see Figure~\ref{fig:pipeline}).
The compression results have been integrated with the bubble data set and are displayed in Figure~\ref{fig:comp_ratio}.
The images of the Kvasir-capsule have a mean CR of $5.578$ $(82.1\%)$ with a PSNR of $\qty{40.26}{\dB}$.
Galar reached a CR of $5.748$ $(82.6\%)$ with a PSNR of $\qty{40.30}{\dB}$.
A notable observation is the inverse relationship between the CR and the bubble coverage, wherein the CR exhibits a substantial decrease as the coverage increases.
When the CR is less than $4$, the images are predominantly obscured by bubbles, rendering it medically irrelevant.
Given that the coverage persists across multiple frames, the fps can be reduced during this interval to conserve power.

\subsubsection{View Labels}
\label{subsec:view_labels}
\begin{figure}[t]
    \centering
    \subfloat[Kvasir-Capsule]{\includegraphics[width=.5\linewidth]{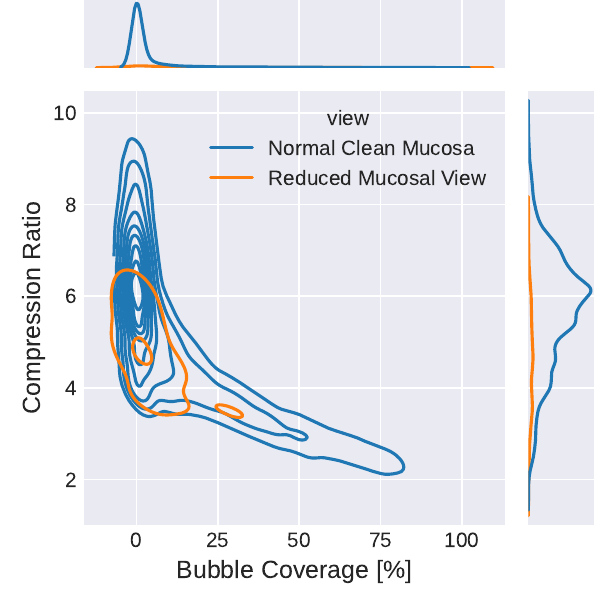}\label{fig:kvasir_view_labels}}
    \subfloat[Galar]{\includegraphics[width=.5\linewidth]{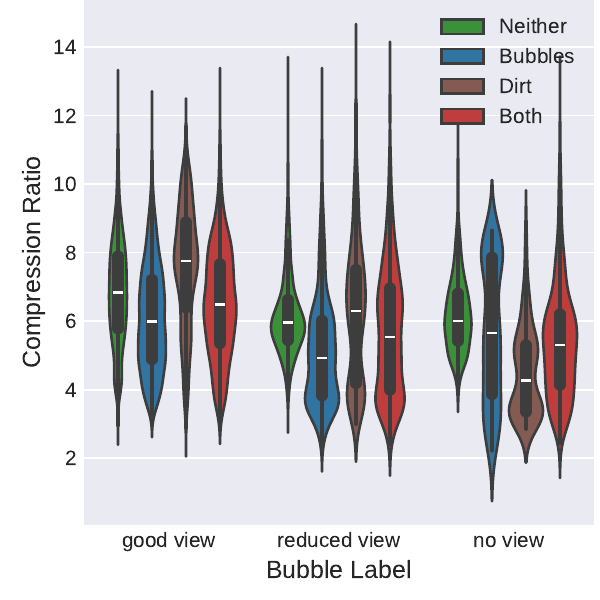}\label{fig:galar_view_labels}}
    \caption{(a) Comparision of two Kvasir-Capsule's luminal finding groups (\textit{normal clean mucosa} and \textit{reduced mucosal view}) and (b) the CR of Galar's technical label group.}
    \label{fig:view_labels}
\end{figure}
\begin{figure}[t]
    \centering
    \subfloat[Normal\\CR: $1.796$]{\includegraphics[width=.25\linewidth]{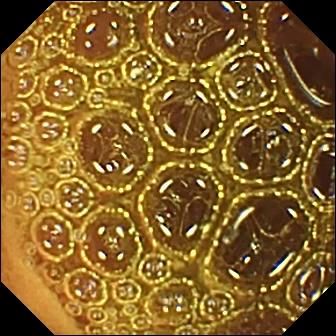}}
    \subfloat[Normal\\CR: $2.755$]{\includegraphics[width=.25\linewidth]{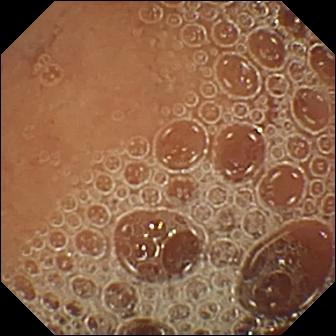}}
    \subfloat[Normal\\CR: $2.778$]{\includegraphics[width=.25\linewidth]{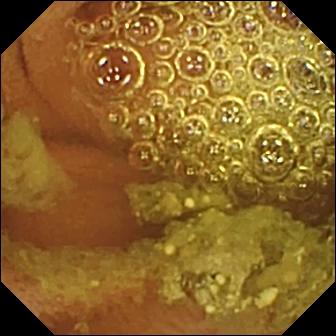}}
    \subfloat[Reduced View\\CR: $7.460$]{\includegraphics[width=.25\linewidth]{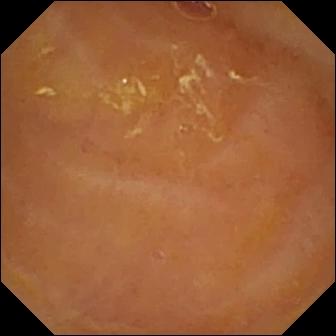}}
    \caption{Inconsistent frame classification of Kvasir-Capsule with \textit{Normal Clean Mucosa} (a-c) and \textit{Reduced Mucosal View} (d) including the frame's achieved CR.}
    \label{fig:bad_kvasir}
\end{figure}

As mentioned in Section~\ref{subsec:datasets}, both datasets offer limited labeling in regard to the visibility of selected frames.
These were used as baseline to verify the correlation between the CR and bubble coverage.
The results are shown in Figure~\ref{fig:view_labels}.
Kvasir-Capsule has $2,906$ images classified as \textit{Reduced Mucosal View}, but did not include a distinction why the view is reduced, and $34,338$ images classified as \textit{Normal Clean Mucosa}, by far the largest group among all labels.
As illustrated in Figure~\ref{fig:kvasir_view_labels}, the first group is harder to compress on average.
However, the normal group has 1,035 images where the achieved CR is below 3.
Four samples that highlight the inconsistent labeling are displayed in Figure~\ref{fig:bad_kvasir}.

Galar's technical labeling strongly supports the correlation thesis, as shown in Figure~\ref{fig:galar_view_labels}.
When the view is constrained by bubbles, the CR sustains the greatest impact.
The restriction of view by dirt has been demonstrated to decrease the achievable CR to a significantly lesser extent than bubbles.
Nevertheless, this decrease remains substantial enough to enable the dropping of them with a high degree of confidence.
It is worth noting that they are also irrelevant to the screening process. 
Dropping these as well constitutes an additional benefit in regard of power consumption.

\subsubsection{Pathologies}
\begin{figure}[t]
    \centering
    \includegraphics[width=\linewidth]{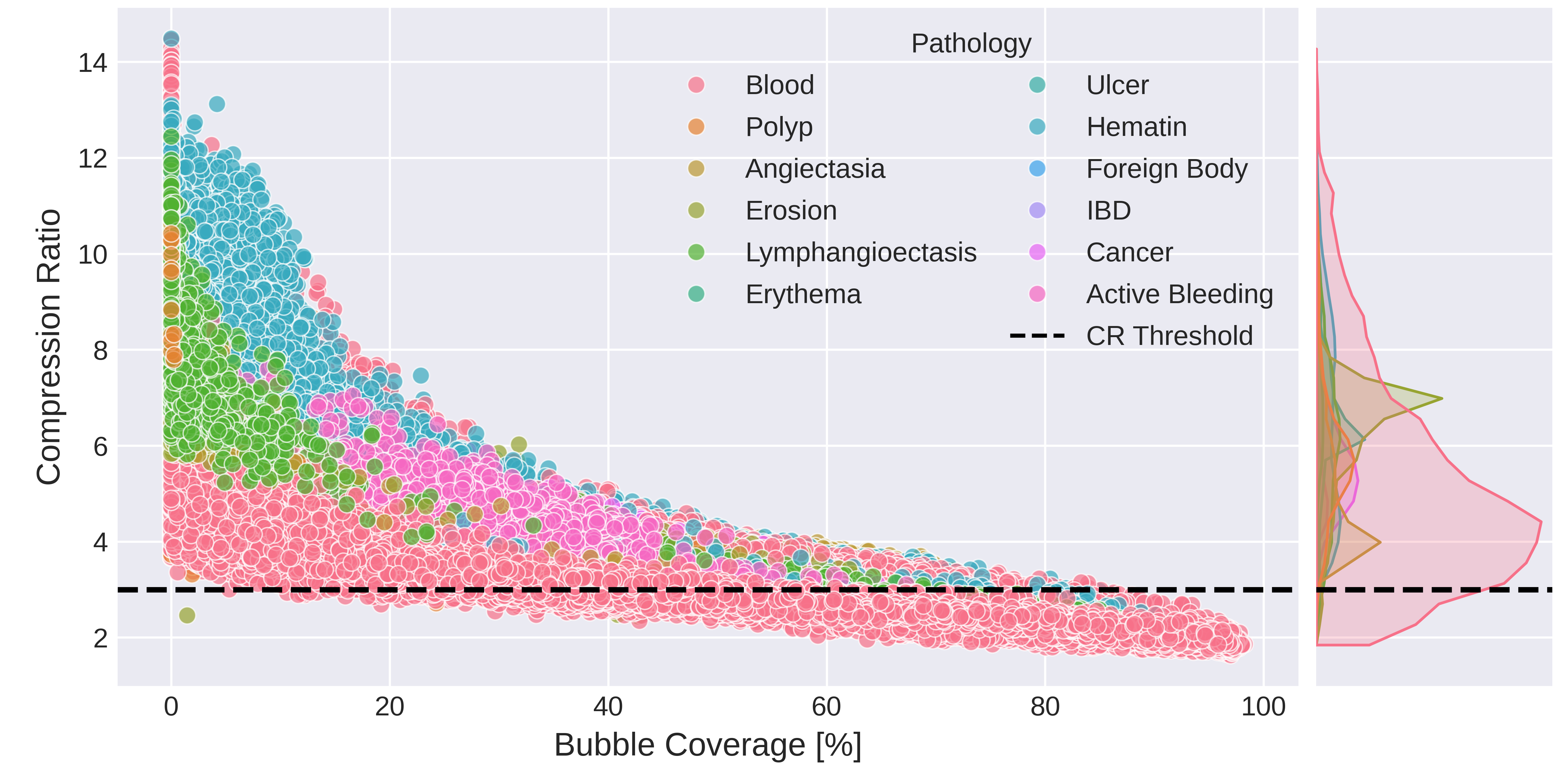}
    \caption{Distribution of the images with a pathology label of the Galar dataset.}
    \label{fig:galar_pathologies}
\end{figure}
Reducing the fps could lead to missed pathologies when they were only visible during the timeframe the capsule did not capture an image.
However, pathologies could also be missed if they were covered by dirt and/or bubbles.
Nevertheless, an analysis of the images with at least one pathology labeled in the small intestine of the Galar dataset revealed, that these frames are mostly compressible with a high CR (see Figure~\ref{fig:galar_pathologies}).
If a pathology can not be compressed efficiently, they were mostly frames with either \textit{Blood} or \textit{Hematin}.
This is due to two factors:
First, $90,735$ and $13,406$ out of all $206,012$ small intestine frames, that were classified with at least one pathology have the label \textit{Blood} and \textit{Hematin}, respectively.
Thus, they are with \textit{Erosion} the largest label groups in the dataset.
Secondly, if these occurrences come together with bubbles, the reflection of the capsule's LEDs have a greater impact on the image quality and its monotonicity, limiting the possibility to compress them effectively.
However, as shown in Section~\ref{subsec:study_sim}, these pathologies occur over an extended period of time, thus they are captured even with a reduced frame rate.

\subsection{Hardware Demonstrator}
\begin{table}[t]
\centering
\setlength{\tabcolsep}{0.5em}
\caption{Power and energy consumption of the demonstrator's modules}
\begin{tabular}{ccccc}
\toprule
Task & Module & IT [$\unit{\ms}$] & Power [$\unit{\mW}$] & Energy [$\unit{\uJ}$] \\
\midrule
\multirow{3}{*}{\begin{tabular}{c}
Image\\
Capture
\end{tabular}} & NanEyeC & $12.79$ & $8.51$ & $108.93$ \\
& LEDs & $12.79$ & $14.78$ & $189.15$ \\
& Core & $12.79$ & $1.06$ & $13.56$ \\
\midrule
Compression & Core & $51.8-94.7$ & $1.06$ & $59-101$ \\
\midrule
Transmission & Transceiver & $7.5-25$ & $5$ & $37.5-125$\\
\midrule
Idle & System & - & $0.43$ & -\\
\bottomrule
\end{tabular}
\label{tab:demonstrator}
\end{table}

\begin{figure}[t]
    \centering
    \subfloat[Relationship between the runtime of the compression pipeline and the achieved CR.]{\includegraphics[width=.48\linewidth]{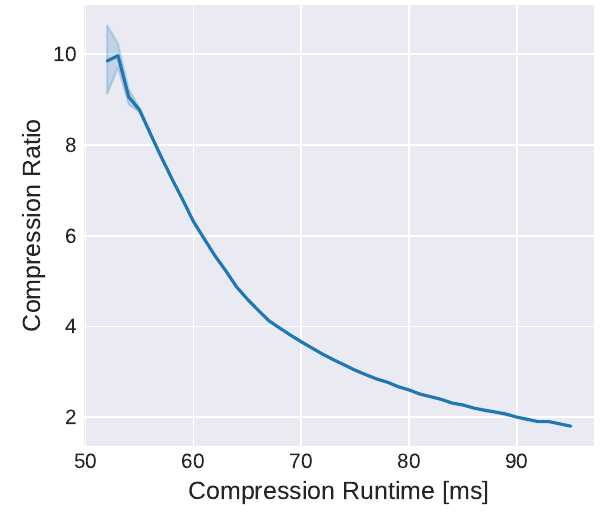}\label{fig:comp_runtime}}
    \hspace{.01\linewidth}
    \subfloat[Number of pathologies skipped by frame rate reduction and CR Threshold combination.]{\includegraphics[width=.48\linewidth,trim={0.5cm 0 0 1.2cm},clip]{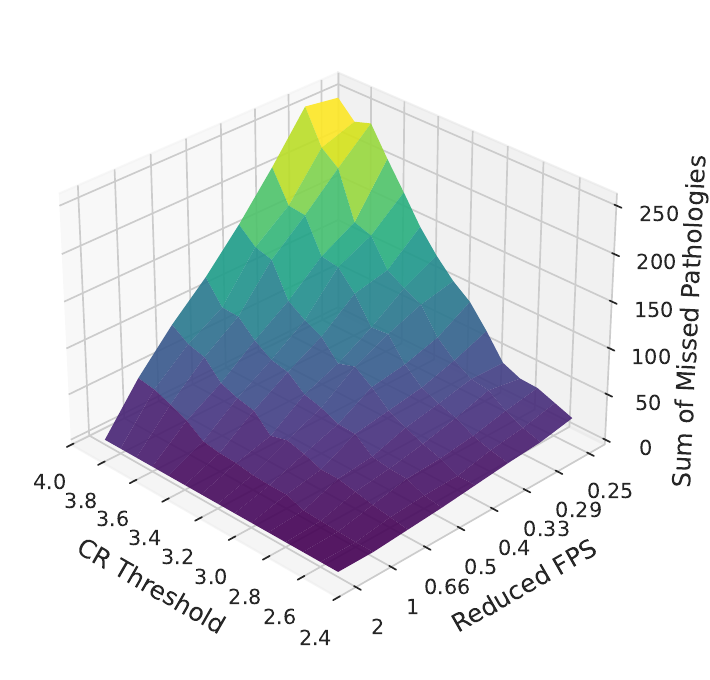}\label{fig:skipped_pathologies}}
    \caption{Demonstrator's runtime and accuracy analysis.}
    \label{fig:xx}
\end{figure}

\begin{table}[t]
\centering
\caption{Performance comparison of different VCE compression pipelines}
\setlength{\tabcolsep}{0.5em}
\begin{tabular}{cccccc}
\toprule
\multirow{2}{*}{Work} & \multirow{2}{*}{CR} & \multirow{2}{*}{PSNR [$\unit{\dB}$]} & \multirow{2}{*}{IT [$\unit{\ms}$]} & Energy per & \multirow{2}{*}{Plattform} \\
 &  &  &  & frame [$\unit{\uJ}$] & \\
\midrule
\cite{mostafa2014improved} & $6.84$ & $52.9$ & - & - & - \\
\cite{lin2006ultra} & $4.91$ & $32.5$ & $<500$ & $7,460$ & HW ACC \\
\cite{khan2011lossless} & $3.57$ & $\infty$ & $<1$ & $143.5$ & HW ACC \\
\cite{turcza2013hardware} & $11.41$ & $35.7$ & $8.2$ & $290$ & FPGA \\
\midrule
Ours & $5.79$ & $40.3$ & $62.8$ & $66.57$ & RISC-V \\
\bottomrule
\end{tabular}
\label{tab:hw_performance}
\end{table}
The demonstrator was equipped with the $320\times320$px NanEyeC image sensor, four LEDs, a transceiver module and a PULPissimo SoC with a single-core RISC-V~\cite{bause2025smartVCE, bernardo2024scalable}.
The SoC has a die size of $1.8\times2\unit{\mm}^2$ and was clocked at $\qty{200}{\MHz}$.
The power and energy consumption of each module are listed in Table~\ref{tab:demonstrator}.
As shown, the capturing of images is by far the most energy-intensive task, followed by the compression and transmission task which costs are highly dependent on the achieved CR of a frame (see Figure~\ref{fig:comp_runtime}).

Table~\ref{tab:hw_performance} summarizes our results on the demonstrator in comparison to literature results. For CR with 5.79, our method achieves comparable results to the other methods while outperforming \cite{lin2006ultra} and \cite{turcza2013hardware} in PSNR with \SI{40.3}{\dB}. 
For \cite{mostafa2014improved} a higher compression and PSNR can be observed; however, no information about inference time and energy consumption are available. 
\cite{khan2011lossless} shows a lossless compression, but the CR (3.57) is about \SI{36}{\percent} lower compared to our method. 
For the inference time, all methods can be considered as real-time capable; however, our method only takes \SI{62.8}{\milli\second} which is eight times lower than~\cite{lin2006ultra}. 
While maintaining sufficient image quality in combination with real-time capability, our method outperforms all other methods in terms of energy per frame which is considered as most important. 
With only \SI{66.57}{\micro\joule}, we achieve a reduction of over \SI{50}{\percent} compared to \cite{khan2011lossless} and about \SI{99}{\percent} compared to \cite{lin2006ultra}.
Additionally, our method can be efficiently executed on any low-power microcontroller while \cite{lin2006ultra} and \cite{khan2011lossless} utilize custom accelerators. Particularly for \cite{khan2011lossless}, the very low inference time of \SI{1}{\milli\second} and the rather low energy consumption can be traced back to this custom HW accelerator.

\subsubsection{Study Simulation}
\label{subsec:study_sim}
\begin{figure*}
    \centering
    \includegraphics[width=\textwidth]{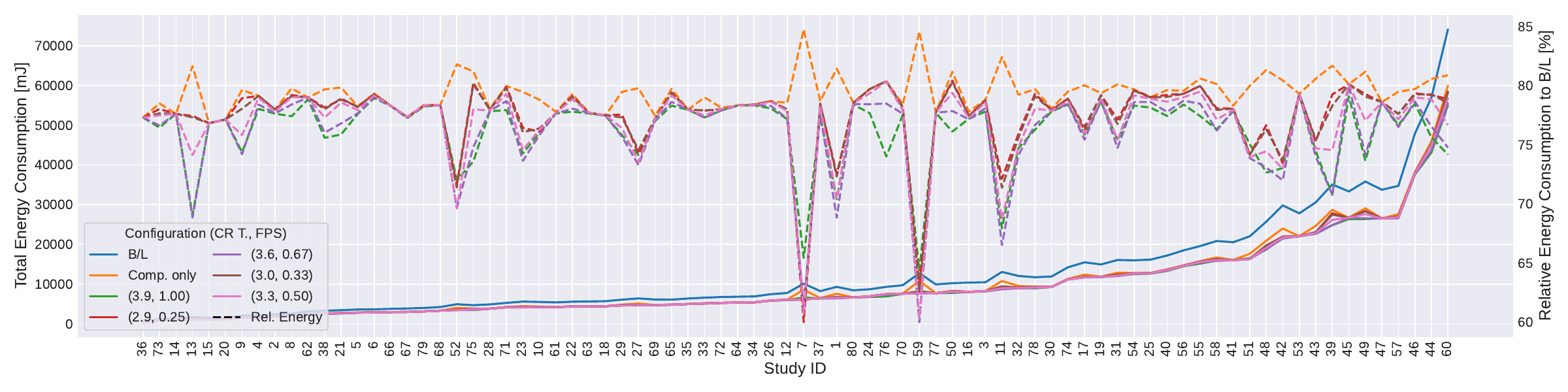}
    \caption{Simulation of energy consumption of the 80 Galar studies with different CR Threshold and fps reduction combinations.}
    \label{fig:hw_power_sim}
\end{figure*}
              
To verify the functionality of our demonstrator and evaluate its performance, the small intestine section of all studies from the Galar dataset were emulated.
For sake of simplicity and comparability, we assume that the frame-rate during all studies was constant at 2 fps.
First, the runtime of the compression pipeline was measured with the over $47,000$ labeled images of Kvasir-Capsule.
The relationship between the runtime and the CR, illustrated in Figure~\ref{fig:comp_runtime}, shows little deviation and, thus, can be modeled by a monotonically decreasing exponential function.
Besides the compression, the energy consumption should also be optimized by temporarily reducing the fps if the capsule's view is severely limited by bubbles.
As shown in Section~\ref{subsec:view_labels}, bubbles decrease the performance of image compression.
However, a suitable CR Threshold and fps reduction needs to be selected in order to prevent missing pathologies.
Figure~\ref{fig:skipped_pathologies} displays the impact of these setting combinations on the total number of missed pathologies across all 80 studies and $200,000$ labeled pathologies.
Obviously, a high CR threshold combined with an aggressive fps reduction leads to a high number of missed pathologies, which needs to be avoided.

The energy simulation results of Galar's 80 studies with different viable configurations are shown in Figure~\ref{fig:hw_power_sim}.
The compression alone reduces the capsule's energy consumption on average by $20.58\%$ and maximum of $23.15\%$ at study $15$.
Note, that the image capturing, which is the largest energy consumer in the capsule, is included in this simulation.
Studies that have many badly compressible frames (upwards spikes in the \textit{comp. only} line) benefit greatly from a temporarily reduced fps.
On average across the settings, the energy consumption was reduced by $23.12\%$.
The greatest reduction was observed at study $59$ with $40\%$ with a CR Threshold of $3.6$ and a reduced fps of $0.67$.
\section{Conclusion}

In this work, we present a bubble-aware, ultra-low-power, hardware-efficient compression pipeline that can be executed on any microcontroller and does not rely on expensive custom accelerators or hardware operators.
The compression pipeline compresses RAW Bayer images, reducing the energy required to transform the image into RGB space. 
The compression algorithm's internal characteristics allow for the parallel detection of bubbles that limit the image's medical significance. 
Using this approach, the frame rate is dynamically adapted with respect to the properties of potential pathologies to avoid missing them. 
Using the proposed compression, compression of over \SI{90}{\percent} was achieved while maintaining high image quality, leading to an average reduction in energy consumption of \SI{20.58}{\percent}. 
Including our dynamic bubble-aware frame rate adaptation achieved an additional reduction in energy consumption of \SI{10.99}{\percent}, without missing any pathologies.
For future research, we will consider further improvements to the compression algorithm by incorporating instruction set extensions that offer greater energy efficiency and a lower inference time. 
Additionally, we intend to enhance the bubble labels for the datasets and make them publicly available.

\bibliographystyle{IEEEtran}
\scriptsize
\bibliography{paper.bib}

\end{document}